\documentclass[sigconf]{acmart}
\makeatletter
\def\@ACM@checkaffil{
    \if@ACM@instpresent\else
    \ClassWarningNoLine{\@classname}{No institution present for an affiliation}%
    \fi
    \if@ACM@citypresent\else
    \ClassWarningNoLine{\@classname}{No city present for an affiliation}%
    \fi
    \if@ACM@countrypresent\else
        \ClassWarningNoLine{\@classname}{No country present for an affiliation}%
    \fi
}
\makeatother

\AtBeginDocument{%
  \providecommand\BibTeX{{%
    \normalfont B\kern-0.5em{\scshape i\kern-0.25em b}\kern-0.8em\TeX}}}

\acmISBN{978-1-4503-XXXX-X/18/06}

\usepackage{geometry}
\usepackage{lipsum}
\usepackage{multicol}
\usepackage{graphicx}
\usepackage{geometry}
\usepackage{array}

\begin{document}
\settopmatter{printacmref=false} 
\renewcommand\footnotetextcopyrightpermission[1]{} 
\pagestyle{plain} 

\title{FACTS About Building Retrieval Augmented Generation-based Chatbots}

   \author{Rama Akkiraju, Anbang Xu, Deepak Bora, Tan Yu, Lu An, Vishal Seth, Aaditya Shukla, Pritam Gundecha, Hridhay Mehta, Ashwin Jha, Prithvi Raj, Abhinav Balasubramanian, Murali Maram, Guru Muthusamy, Shivakesh Reddy Annepally, Sidney Knowles, Min Du, Nick Burnett, Sean Javiya, Ashok Marannan, Mamta Kumari, Surbhi Jha, Ethan Dereszenski, Anupam Chakraborty, Subhash Ranjan, Amina Terfai, Anoop Surya, Tracey Mercer, Vinodh Kumar Thanigachalam, Tamar Bar, Sanjana Krishnan, Jasmine Jaksic, Nave Algarici, Jacob Liberman, Joey Conway, Sonu Nayyar and Justin Boitano} 
    \affiliation{ 
      \institution{NVIDIA}
    }
    \email{{rakkiraju, anbangx, dbora}@nvidia.com}
\begin{abstract}
Enterprise chatbots, powered by generative AI, are rapidly emerging as the most explored initial applications of this technology in the industry, aimed at enhancing employee productivity. Retrieval Augmented Generation (RAG), Large Language Models (LLMs), Langchain/Llamaindex types of LLM orchestration frameworks serve as key technological components in building generative-AI based chatbots. However, building successful enterprise chatbots is not easy. They require meticulous engineering of RAG pipelines. This includes fine-tuning semantic embeddings and LLMs, extracting relevant documents from vector databases, rephrasing queries, reranking results, designing effective prompts, honoring document access controls, providing concise responses, including pertinent references, safeguarding personal information, and building agents to orchestrate all these activities. In this paper, we present a framework for building effective RAG-based chatbots based on our first-hand experience of building three chatbots at NVIDIA: chatbots for IT and HR benefits, company financial earnings, and general enterprise content. Our contributions in this paper are three-fold. First, we introduce our FACTS framework for building enterprise-grade RAG-based chatbots that address the challenges mentioned. FACTS mnemonic refers to the five dimensions that RAG-based chatbots must get right - namely content \underline{f}reshness (F), \underline{a}rchitectures(A), \underline{c}ost economics of LLMs  (C), \underline{t}esting (T), and \underline{s}ecurity (S). Second, we present fifteen control points of RAG pipelines and techniques for optimizing chatbots’ performance at each stage. Finally, we present empirical results from our enterprise data on the accuracy-latency tradeoffs between large LLMs vs small LLMs.  To the best of our knowledge, this is the first paper of its kind that provides a holistic view of the factors as well as solutions for building secure enterprise-grade chatbots.

\end{abstract}




\maketitle

\section{Introduction}

Chatbots are increasingly becoming an extension of search tools in companies for finding relevant information. Whether it's HR benefits, IT help, sales queries, or engineering issues, enterprise chatbots are now go-to productivity tools. Before the debut of OpenAI's Chat-GPT~\cite{achiam2023gpt} in November 2022, companies relied on internally developed chatbots  based on dialog flows. Such bots required extensive training for intent understanding and meticulous orchestration for response generation and yet could only provide extractive answers at best. These early bots, built on dialog management systems paired with information retrieval and question answering (IRQA) solutions were fragile and limited in capability. While previous generation language models and GPT models existed, they lacked the accuracy, robustness, and reliability needed for broad enterprise use~\cite{galitsky2019developing}. 

Chat-GPT's release, the emergence of vector databases, and the wide-spread use of retrieval augmented generation (RAGs)~\cite{lewis2020retrieval} marked the beginning of a new era in Chatbot domain. Now, LLMs can understand user intents with simple prompts in natural language, eliminating the need for complex intent variant training, synthesize enterprise content coherently, thereby empowering chatbots with conversational capability beyond scripted intent recognition. While LLMs bring their generative capabilities to construct coherent, factual, and logical responses to user queries, vector database-powered information retrieval (IR) systems augment LLMs ability to retrieve fresh content. Tools like LangChain~\cite{Langchain} and Llamaindex~\cite{Liu_LlamaIndex_2022} facilitate chatbot construction, and orchestration of complex workflows including memory, agents, prompt templates, and overall flow. Together, vector-search based IR systems, LLMs, and LangChain-like frameworks form core components of a RAG pipeline and are powering generative AI chatbots in post Chat-GPT era.

\begin{table*}[tp!]
\small
\caption{ A summary of the three chatbots and the current state of development. }
\begin{tabular}{c|c|c|c|c|c|c}
\hline 
 Chatbot & Domain & {Data Sources}  & {Data Types}  &   {Access Control}  & {Sample Queries}  & {State}  \\ \hline \hline
 \shortstack{NVInfo \\Bot }&  \shortstack{Enterprise Internal\\   Knowledge}  & \shortstack{SharePoint, GoogleDrive, Slack\\  Confluence, ServiceNow, Jira \emph{etc.}}  & \shortstack{Docs, HTML \\ PDFs, Slides} & Yes & \shortstack{Can I park overnight \\ in HQ parking lots?   } &  \shortstack{Early Access \\ Testing} \\ \hline \hline
 \shortstack{NVHelp \\ Bot} &  \shortstack{IT Help \\ HR Benefits} &  \shortstack{Knowledge Articles for ITHelp \\ HR benefits pages } &  \shortstack{Text, PDFs \\ Docs}  & Yes &  \shortstack{How to enroll in Employee \\ Stock Purchase plan?}  &  Production\\ \hline \hline
 \shortstack{Scout \\ Bot}&  Financial Earnings & \shortstack{Company news, blogs, SEC filings \\ Earnings-related interviews }  & \shortstack{HTML, PDFs \\ Docs}  & No & \shortstack{What are NVIDIA revenues \\ for the past 3 years? } & Production \\ \hline
\end{tabular}
\label{tab:summarize}
\end{table*}

At NVIDIA, our main motivation was to improve our employee productivity by building enterprise chatbots. Our initial enthusiasm quickly met with the reality of addressing numerous challenges. We learned that crafting a successful enterprise chatbot, even in post Chat-GPT era, while promising, is not easy. The process demands meticulous engineering of RAG pipelines, fine-tuning LLMs, and engineering prompts, ensuring relevancy and accuracy of enterprise knowledge, honoring document access control permissions, providing concise responses, including pertinent references, and safeguarding personal information. All of these require careful design, skillful execution, and thorough evaluation, demanding many iterations. Additionally, maintaining user engagement while optimizing for speed and cost-efficiency is essential. Through our journey, we learned that getting an enterprise conversational virtual assistant right is akin to achieving a perfect symphony where every note carries significance!

In this paper, we share our experiences and strategies in building effective, secure, and cost-efficient chatbots. We answer the following questions from a practitioner perspective:

\begin{itemize}
\item \textit{What are the key challenges to consider when building and deploying enterprise-grade generative AI-based chatbots?} We present our findings from trying to deliver \textit{\underline{f}}resh content (F) with flexible \textit{\underline{a}}rchitectures (A) that are \textit{\underline{c}}ost-efficient (C), \textit{\underline{t}}ested well (T), and \textit{\underline{s}}ecure (S) - (FACTS).  
\item \textit{How to achieve user acceptable levels of quality with RAG systems in building chatbots?} We present the fifteen control points of RAG pipelines and techniques for optimizing each control point and the overall RAG pipeline.
\end{itemize}


\section{Case Study}

Our company's content landscape includes both authoritative knowledge and unauthoritative content. Authoritative content encompasses IT help articles, HR resources in platforms like ServiceNow, and project documentation on Confluence, SharePoint, Google Drive, and engineering tools like NVBugs and GitHub. Employee-generated content complements these sources on platforms such as Slack and MS Teams. In this paper, we present three bots that we have built at NVIDIA using RAGs and LLMs. These bots are briefly introduced below. All three bots are built on our in-house built generative-AI chatbot platform called NVBot platform. Some of the queries that our bots are capable of answering are shown in Table~\ref{tab:summarize}.

\begin{itemize}
    \item \textbf{NVInfo Bot} answers questions about enterprise content (approx. 500M documents of size > 7 TB), complementing intranet search. It manages diverse data formats and enforces document access controls. The tech stack includes LangChain, a vendor vector database for retrieval and to handle document access controls, LLM model (multiple LLM models can be selected), and a custom web-UI.
    \item \textbf{NVHelp Bot} Bot focuses on IT help and HR benefits (approx. 2K multi-modal documents containing text, tables, images, pdfs, and html pages), using a similar tech stack to NVInfo bot with a smaller data volume.
    \item \textbf{Scout Bot} handles questions about financial earnings from public sources, managing structured and unstructured data (approx. 4K multi-modal documents containing text, tables, pdfs, and html pages). The tech stack includes an Open source Vector DB, LangChain, Ragas evaluation, selectable LLM models, and a custom web-UI.
\end{itemize}

In the remainder of the paper, we present our FACTS framework that summarizes the challenges experienced and the learnings gained in building the aforementioned three chatbots. We first start with the challenge of dealing with delivering fresh enterprise content in each of the chatbots.

\begin{figure*}[h]
  \includegraphics[width=0.8\textwidth]{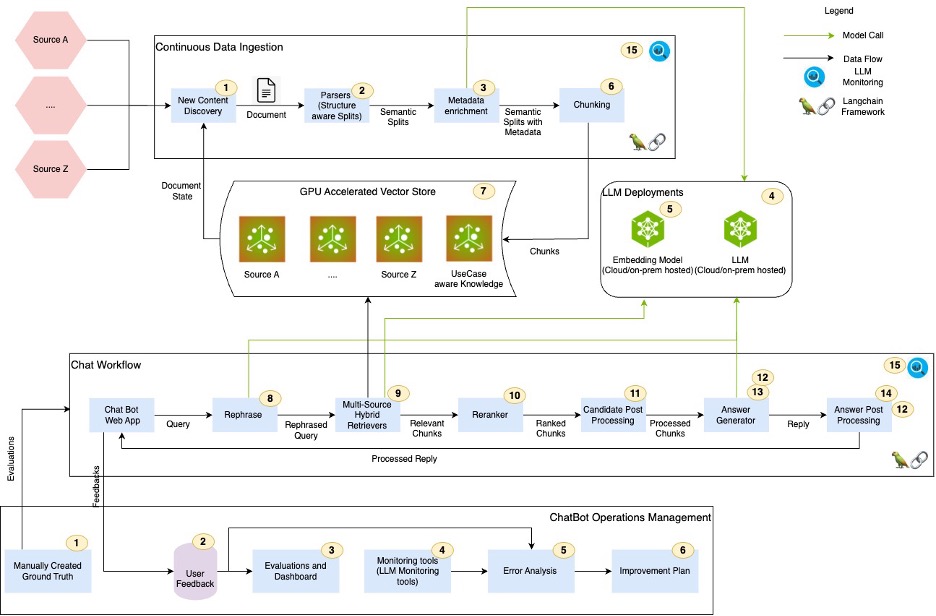}
  \caption{Control Points in a typical RAG pipeline when building Chatbots.}
  \label{fig:RAG_control_points}
\end{figure*}

\section{Ensuring Freshness of Enterprise Data in LLM-powered Chatbots (F)}

Ensuring the freshness of enterprise data in LLM-powered chatbots presents several challenges. Foundation models, although powerful, often fall short as they lack domain-specific and enterprise-specific knowledge. Once trained, these models are essentially frozen in time and may hallucinate, providing undesired or inaccurate information when used on enterprise content that they are not trained on. 

Retrieval Augmented Generation (RAG) is a process where relevant information is retrieved from vector databases through semantic matching and then fed to LLMs for response generation. In a RAG pipeline, vector databases and LLMs collaboratively ensure the delivery of up-to-date enterprise knowledge. However, RAG pipelines have many control points, each of which when not tuned well can lead to lower accuracy, hallucinations, and irrelevant responses by Chatbots. Additionally, document access control permissions complicate the search and retrieval process, requiring careful management to ensure data security and relevance. Furthermore, multi-modal content necessitates the use of multi-modal retrievers to handle structured, unstructured, and semi-structured data, including presentations, diagrams, videos, and meeting recordings. Addressing these challenges is critical for maintaining the accuracy and reliability of enterprise chatbots. Inspired by \cite{barnett2024seven}, we identify fifteen control points of RAG from our case studies visualized in Figure~\ref{fig:RAG_control_points}. Each control point is labeled with a number. In the remainder of this section, we present our insights and learnings for addressing RAG control points.

\subsection{Learnings}
\begin{figure*}[t]
    \centering
    \includegraphics[width=0.8\textwidth]{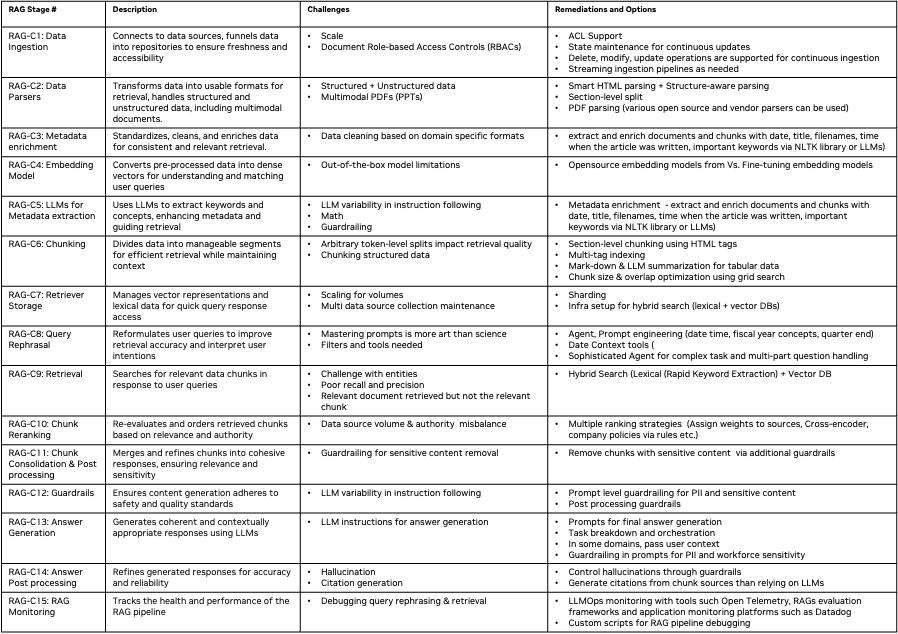}
    \vspace{-3mm}
    \caption{RAG control points, challenges, and remediations}
    \vspace{-3mm}
    \label{fig:RAG_remediations}
\end{figure*}

In figure~\ref{fig:RAG_remediations}, we present a summary description of the fifteen control points of RAG pipelines, challenges associated with each control point, and our suggested approaches for optimizing each control point. Each control point is labeled as RAG-C[num] and RAG-Op[num] for RAG and RAGOps flows, respectively. Below, we present a few key learnings and insights to manage the fresh enterprise content. 

\noindent \textbf{Metadata Enrichment, Chunking, Query Rephrasal, Query Reranking}: We noticed that metadata enrichment, chunking, query rephrasal and query re-ranking stages of RAG pipeline have the most impact on the quality of Chatbot responses. LLM response generation quality is highly dependent on retrieval relevancy. Retrieval relevancy is, in turn, highly dependent on document metadata enrichment, chunking, and query rephrasal. We implemented grid search-based auto-ML capabilities to find the right configurations of chunk token-sizes, experimented with various prompt variations, and explored different chunk reranking strategies to find optimal settings for each. While we have made significant improvements in retrieval relevancy and answer quality and accuracy, 
we believe, we still have more work to do to optimize the full pipeline.

\noindent \textbf{Hybrid Search}: We noticed that Vector databases are not so good at handling matching entities (e.g., people names, places, company names etc.). Using a combination of Lexical search (e.g., elastic search) and vector search provided better retrieval relevancy and more coverage. Setting up an infrastructure that supports hybrid search capabilities, which combines the strengths of both lexical and vector-based searches, can enhance the accuracy and speed of the retrieval process.

\begin{figure}[tp!]
    \centering
    \includegraphics[width=0.45\textwidth]{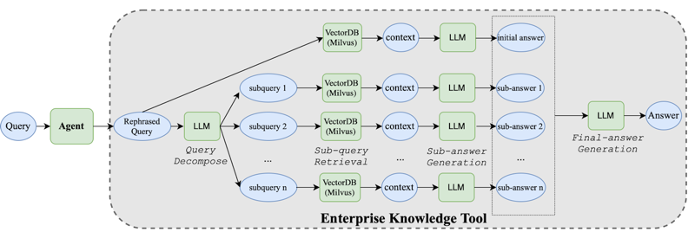}
    \vspace{-3mm}
    \caption{Agent architecture for handling complex queries}
    \vspace{-3mm}
    \label{fig:complx_agent_arch}
\end{figure}

\noindent \textbf{Agentic Architectures}:  Questions such as ‘compare the revenue of NVIDIA from Q1 through Q4 of FY2024 and provide an analytical commentary on the key contributing factors that led to the changes in revenues during this time’ require complex agents that are capable of query decomposition and orchestration. Figure~\ref{fig:complx_agent_arch} shows one mechanism we had implemented to deal with such questions in Scout bot. From our experience of building the three bots, we have realized that IR systems and LLMs are insufficient for answering complex queries. Complex agents and multi-agent architectures are needed to handle complex queries.

\begin{figure}[tp!]
    \centering
    \includegraphics[width=0.45\textwidth]{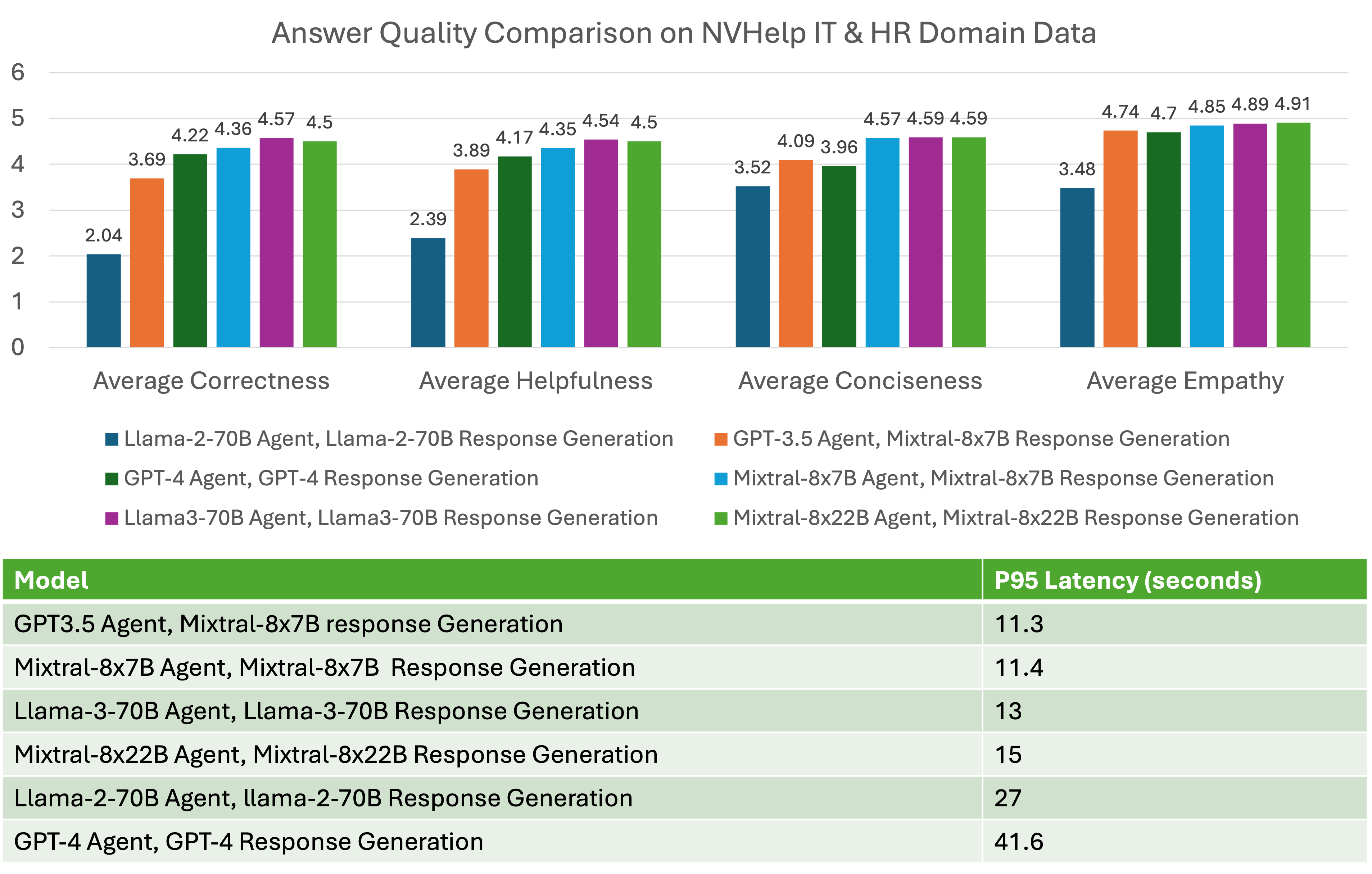}
    \vspace{-3mm}
    \caption{NVHelp answer quality and latency metrics comparison among different models}
    \vspace{-3mm}
    \label{fig:modelperf}
\end{figure}

\noindent \textbf{To Fine-tune LLMs or not?} A key decision is whether to fine-tune LLMs, balancing the use of foundational models with domain-specific customizations. One size doesn’t fit all when it comes to LLMs. Some use cases may work well with foundational models, while others require customization. When considering customization, several options are available, including prompt engineering, P-tuning, parameter-efficient fine-tuning (PEFT), and full fine-tuning (FT). Fine-tuning requires significant investment in data labeling, training, and evaluations, each of which can be time-consuming and costly. Automating testing and quality evaluation processes become critical to ensuring efficiency and accuracy when customizing LLMs. Figure~\ref{fig:modelperf} shows the accuracy vs latency tradeoff evaluations we have done comparing OpenAI’s GPT-4 model with some of the open-source models on about 245 queries from NVHelp bot domain. Our results show that the Llama3-70B model excels in several aspects of answer quality while maintaining acceptable latency.

\noindent \textbf{Handling multi-modal data}: Enterprise data is multi-modal. Handling structured, unstructured, and multi-modal data is crucial for a versatile RAG pipeline. From our experience, if the structure of the document is consistent and known apriori (like those found in EDGAR databases for SEC filings data in financial earnings domain that Scout bot was handling), implementing section-level splitting, using the section titles and subheadings and incoporating those in the context of chunks improves retrieval relevancy. We also found solutions like Unstructured.io, which specialize in extracting and structuring content from PDFs, helpful in parsing and chunking unstructured documents with context.

\noindent \textbf{RAGOps}: Effective health monitoring of RAG pipelines is essential once they are deployed. When answer quality is poor, a thorough error analysis is required to determine whether the issue lies in retrieval relevancy or LLM response generation. To debug retrieval relevancy, developers need detailed information on which chunks were stored in vector databases with their associated metadata, how queries were rephrased, which chunks were retrieved, and how those chunks were ranked. Similarly, if an LLM response is incorrect, it is crucial to review the final prompt used for answer generation. For issues with citations, developers must trace back to the original document links and their corresponding chunks. RAGOps/LLMOps and evaluation frameworks, such as Ragas, are critical for providing the necessary automation to enable rapid iteration during accuracy improvement cycles in RAG pipelines. 

More details on each control point are presented in Figure~\ref{fig:RAG_remediations}. In summary, while promising, implementing RAG systems for chatbots demands meticulous planning and continuous evaluation to ensure secure and accurate data retrieval.

\begin{figure}[t]
  \includegraphics[width=0.45\textwidth]{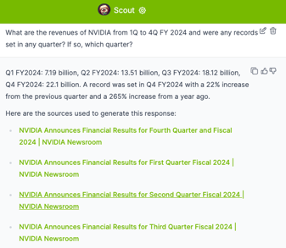}
  \vspace{-3mm}
  \caption{Scout Bot: Multi-part query}
  \label{fig:scout_multipart_query}
  \vspace{-3mm}
\end{figure}

\begin{figure}[h]
  \includegraphics[width=0.45\textwidth]{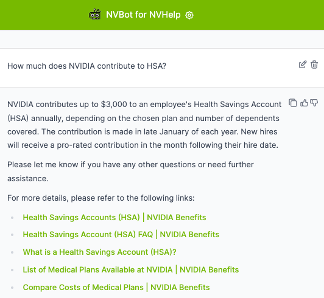}
  \vspace{-3mm}
  \caption{NVHelp Bot: Answering questions on HR benefits}
  \vspace{-3mm}
  \label{fig:nvhelp_bot}
\end{figure}

\vspace{-3mm}
\section{Building Flexible Architectures for generative AI chatbots (A)}

Keeping up with rapid progress in AI is like navigating a fast-moving river. Every aspect, from vector databases and embedding models to LLMs, agentic architectures, low-code/no-code platforms, RAG evaluation frameworks, and prompting techniques, is evolving rapidly. Concurrently, departments within companies are exploring generative AI by building their own chatbots and AI copilots.

In this dynamic environment, building common, flexible, and adaptive platforms are crucial. At NVIDIA, our chatbot ecosystem has grown significantly, reflecting a trend likely seen in many companies. From building three initial chatbots, we realized the importance of a common platform to avoid duplicated efforts in security, guardrails, authentication, prompts, user interfaces, feedback mechanisms, usage reporting, monitoring, and evaluations.

\begin{figure*}[ht]
    \centering
    \includegraphics[width=0.8\textwidth]{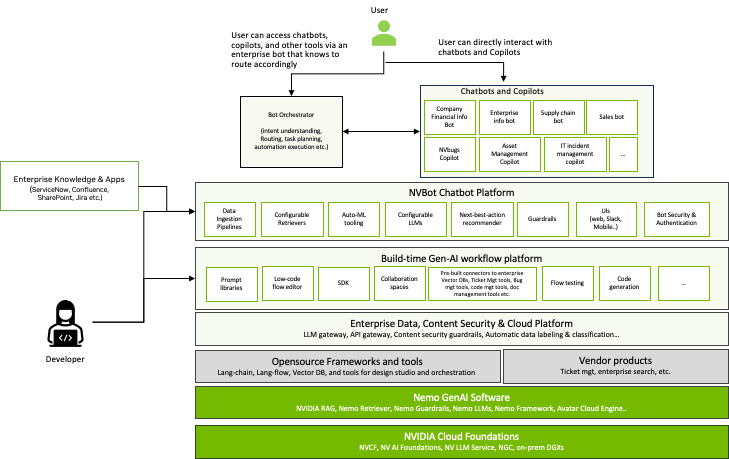}
     \vspace{-1mm}
    \caption{Architecture of NVBot platform upon which multiple chatbots are being built.}
     \vspace{-2mm}
    \label{fig:arch}
\end{figure*}

To address this, we developed the NVBot platform (Figure~\ref{fig:arch}), a modular platform with a pluggable architecture. It allows developers to select LLMs, vector databases, embedding models, agents, and RAG evaluation frameworks that best suit their use case. It also provides common components for essential features like security, guardrails, authentication, authorization, user experience, and monitoring. Additionally, the platform supports citizen development, allowing multiple teams to contribute their tested prompts, workflows, guardrails, and fine-tuned models for collective use.

As our ecosystem of bots expanded, we faced a critical question: should organizations build many domain-specific bots, a single enterprise bot, or go with a hybrid approach? Domain-specific chatbots excel in tailored environments, while nterprise-wide chatbots act as generalists, providing a centralized knowledge base for all employees. Through our experience, we realized that there is no need to choose one over the other.

Novel architectural patterns are emerging where enterprise-wide chatbots act as `switchboards', directing inquiries to specialized bots tuned with domain-specific data. This multibot architecture allows for the concurrent development of specialized chatbots while providing users with a unified interface. Our NVBot platform supports the coexistence and orchestration of multiple chatbots within an enterprise. The debate over a single bot or multiple specialized bots is ongoing. We envision a landscape where domain-specific bots coexist with a centralized information bot, supported by 'copilots'—generative AI capabilities integrated into workplace environments like programming IDEs and collaboration tools. At NVIDIA, we're actively exploring all three chatbot variations—domain-specific, enterprise-wide, and copilot as generative AI reshapes workplace efficiency and information accessibility.

\section {Cost Economics of Chatbot deployments (C)}

Understanding the cost economics of generative AI-based chatbots involves several critical factors. The high costs of major and commercial LLMs can be unsustainable, with expenses adding up significantly across multiple use cases. Additionally, unseen expenses often accumulate as teams test various LLMs to meet specific needs. Moreover, when using commercial LLM vendor APIs, securing sensitive enterprise data requires guardrails to detect and prevent sensitive data leakage, as well as gateways for audit and legally permitted learning. There are also cost versus latency trade-offs to consider, as large LLMs with long context lengths typically have slower response times, impacting overall efficiency.

\textbf{Bigger Vs. Smaller Models}: Larger, commercial LLMs, smaller open source LLMs are increasingly becoming viable for many use cases, thereby offering cost effective alternatives to companies. As opensource models are catching up with larger, commercial models, they are increasingly offering close-comparable accuracy, as demonstrated in our NVHelp bot emperical evaluation in Figure~\ref{fig:modelperf}, and generally have better latency performance compared to larger models. Additionally, GPU optimization of inference models can further speed up processing times. Open-source models optimized with NVIDIA’s Tensor RT-LLM inference libraries, for instance, have shown faster performance than non-optimized models. These strategies help balance the need for cost-efficiency with maintaining high performance and security standards.

\textbf{LLM Gateway}: If you must use a vendor LLM API, it is better to implement an internal company LLM Gateway for audit, subscription and cost management across the company. Implementing an internal company LLM Gateway can streamline LLM usage, subscriptions, and data tracking for security audits. This central hub simplifies management and ensures efficient resource allocation. At NVIDIA IT, we have implemented an LLM Gateway that logs the inbound and outbound payloads for audit purposes and this data is guarded with access control permissions. Our LLM Gateway helps manage the subscriptions and costs of LLM API invocations. 

In summary, developing a hybrid and balanced LLM strategy is essential for managing costs and enabling innovation. This involves using smaller and customized LLMs to manage expenses while allowing responsible exploration with large LLMs via an LLM Gateway. It's crucial to measure and monitor ROI by keeping track of LLM subscriptions and costs, as well as assessing Gen-AI feature usage and productivity enhancements. Ensuring the security of sensitive enterprise data in cloud-based LLM usage requires implementing guardrails to prevent data leakage and building an LLM Gateway for audits and legally permitted learning. Finally, be aware of the trade-offs between cost, accuracy, and latency, customizing smaller LLMs to match the accuracy of larger models while noting that large LLMs with long context lengths tend to have longer response time.

\vspace{-3mm}
\section{Testing RAG-based Chatbots (T)}
Testing generative AI solutions can be a lengthy process due to the need for human response validation. LLMs are increasingly being employed using ‘LLM-as-a-judge’ approach. However, it is advisable to use caution when using LLMs as human proxy, as using LLMs as judges can lead to self-fulfilling prophecy type of scenarios reinforcing their inherent biases in evaluations as well. 
\begin{itemize}
\item{\textbf{Security Testing}}: Automating security testing is critical for maintaining development velocity without compromising safety. A strong security framework and regression test datasets ensure that the chatbot remains resilient to potential threats. We are collaborating with our internal RED teams in security to prepare a set of datasets that can be tested with each major iteration of the chatbot.

\item{\textbf{Prompt Change Testing}}: Generative AI models can be highly sensitive to prompt changes. To maintain accuracy, full regression testing is needed with each prompt alteration. 

\item{\textbf{Feedback Loops}}: Incorporating feedback gathered and the RLHF cycle is pivotal for continuous improvement. It allows LLM models to refine both our solutions and Language Models over time, ensuring that the chatbot becomes increasingly proficient. However, if the chosen foundational models don’t offer customization, then it becomes difficult to align the models to human feedback. If the feedback is significant and comes in many areas, then model customization may be an option. As of now, we have begun gathering user feedback but haven’t built our continuous learning pipelines using RLHF yet. Having tools to make this automated is critical to pos-production life cycle management of these chatbots.

\end{itemize}

\vspace{-1mm}
\subsection{Learnings}

\textbf{Plan for Long Test Cycles}: Effective testing of RAG-based chatbots requires anticipation of lengthy test cycles. Begin by focusing on automating tests and enhancing accuracy assessments to streamline this essential phase.

\noindent {\textbf{Build Representative Ground Truth Datasets}}: It is crucial to construct comprehensive ground truth datasets that reflect full spectrum of targeted solution strengths. This ensures that the chatbot is tested against scenarios that it will encounter in actual use.

\noindent{\textbf{Automate Evaluations}}: While leveraging LLMs as evaluators can provide scalable testing options, remember that the quality of human evaluations is unmatched. Automated tools should be used where feasible to supplement but not replace human oversight.

\noindent {\textbf{Incorporate Human Feedback and Continuous Learning}}: Establish mechanisms that allow for  human feedback and systematic error analysis. Prioritize iterative improvements based on this feedback to continually refine chatbot performance and adaptability.

\vspace{-3mm}
\section {Securing RAG-based Chatbots (S)}

Building trust is paramount when deploying generative AI chatbots. To mitigate risks, guardrails for hallucinations, toxicity, fairness, transparency, and security are critical. Strong foundational models are increasingly getting better at these guardrails. However, there are still many possibilities of jail breaks, adversarial attacks, and other security issues. Apart from these security risks, generative AI-based chatbots are susceptible to derivative risks (explained below). Since our bots are all internal enterprise chatbots, our focus has been more on the enterprise content security and guardrailing for sensitive data. Below we summarize our learnings and insights for securing RAG-based chatbots based on our experience. Addressing these challenges is imperative to maintaining the integrity and security of RAG-based chatbots within corporate environments.

\vspace{-1mm}
\subsection{Learnings}

\noindent {\textbf{Enterprise Content Access Control}}: Enterprise documents are protected by access controls, requiring RAG-based chatbots to comply with Access Control Lists (ACLs) during response generation. To ensure this compliance, we specifically selected an IR product known for its capability to honor these document ACLs effectively. 

\noindent{\textbf{Derivative Risks with Generative AI}}: Chatbots might generate responses that lack context from their original data sources, potentially leading to misinterpretations. Additionally, enhanced search methods could inadvertently elevate the risk of exposing sensitive data if enterprise content is inappropriately secured. As part of our NVInfo bot journey, we implemented sensitive data guardrails in addition to leveraging sensitive data filtering and classification capabilities provided by the vector search solution we used to automatically filter out sensitive data during the retrieval.

\noindent{\textbf{Data Governance and Content Security}}: Efficient knowledge access can increase sensitive data leakage risks. Thus, it's essential to prioritize data governance before deployment to safeguard against unauthorized access and data breaches. At NVIDIA, we embarked on an enterprise content security initiative for  document sensitivity classification and exclusion of sensitive content from chatbots.

\noindent {\textbf{Enterprise Guardrailing}}: Implementing guardrails that align generative AI responses with specific enterprise policies and rules is essential. These guardrails help mitigate risks by ensuring that Chatbot-generated content adheres to established norms and ethical guidelines, preventing potential legal and reputational damage. In NVInfo bot, we implemented many guardrails in LLM prompts initially. However, later realized that not all LLMs follow these prompts consistently. Therefore, we implemented these guardrails during pre and post processing of queries and responses respectively using Nemo Guardrails~\cite{rebedea2023nemo}.

\section{Related work}

Our work can be compared with RAG papers on various topics dealing with RAG quality along all the FACTS dimensions we presented (freshness, architecture, costs, testing and security). Due to lack of space, we contrast our work with selective works. Barnett~\emph{et. al.}~\cite{barnett2024seven} presented seven failure points when engineering RAG systems. In their work, they highlight the challenges of getting retrieval augmented generation right by presenting their findings from having built three chatbots. Wenqi Glantz~\cite{wq2024} elaborated 12 RAG pain points and presented solutions. We experienced similar challenges first-hand when building our chatbots. However, none of these works discuss the challenges with complex queries, testing, dealing with document security, and the need for flexible architectures. In our work, we not only build on failure/pain points of RAGs as mentioned above, but also present our 15 control points in RAG pipelines and offer specific solutions for each stage. Also, we extend our insights and present practical techniques for handling complex queries, testing, and security. We present a reference architecture for one of the implementations of agentic architectures for complex query handling, strategies for testing and evaluating subjective query responses, and raised awareness for dealing with document ACLs and security. Furthermore, we present a reference architecture for a flexible generative-AI based Chatbot platform. 

ChipNemo~\cite{liu2023chipnemo}  presents evidence for using a domain adapted language model for improving RAG’s performance on domain specific questions. They finetuned the e5-small-unsupervised model with 3,000 domain specific auto-generated samples. We tried fine-tuning e5-large embeddings model in Scout Bot. Our results did not demonstrate significant improvements. We are presently collecting high quality human-annotated data to repeat the experiments. This could be an important direction to explore in the future for our work. Another interesting technique was presented by Setty~\emph{et. al.}~\cite{setty2024improving}, in improving RAG performance using Hypothetical Document Embeddings (HYDE) technique.  HyDE uses an LLM to generate a theoretical document when responding to a query and then does the similarity search with both the original question and hypothetical answer. This is a promising approach  but might make the architecture complex. 

Active Retrieval augmented generation (FLARE)~\cite{jiang2023active} iteratively synthesizes a hypothetical next sentence. If the generated sentence  contains low-probability tokens,  FLARE would use the sentence as the new query for retrieval and regenerate the sentence. Mialon~\emph{et al.}~\cite{mialon2023augmented} reviews works for advanced augmented generation methods in language model. Self-refine \cite{madaan2024self} builds an agent to improve the initial answer of RAG through iterative feedback and refinement. ReAct~\cite{yao2022react} Agent is widely used for handling the complex queries in a recursive manner.  On the RAG evaluation front, RAGAS~\cite{es2023ragas} and ARES~\cite{saad2023ares} utilize LLMs as judges and build automatic RAG benchmark to evaluate the RAG system.  Zhu~\emph{et al.}~\cite{zhu2023large} overview the intensive usages of LLM in a RAG pipeline including retriever, data generation, rewriter, and reader. We believe that our work provides a unique perspective on building secure enterprise-grade chatbots via our FACTS framework. 
 
\section{Conclusions}
In this paper, we presented our approach to developing effective RAG-based chatbots, highlighting our experiences of building three chatbots at NVIDIA. We outlined our FACTS framework, emphasizing the importance of content freshness (F), architecture (A), LLM cost (C) management, planning for testing (T), and security (S) in creating robust, secure, and enterprise-grade chatbots. We also identified and elaborated on fifteen critical control points within RAG pipelines, providing strategies to enhance chatbot performance at each stage. Furthermore, our empirical analysis reveals the trade-offs between accuracy and latency when comparing large and small LLMs. This paper offers a holistic perspective on the essential factors and practical solutions for building secure and efficient enterprise-grade chatbots, making a unique contribution to the field. 
More work is needed in several areas to build effective RAG-based chatbots. This includes developing agentic architectures for handling complex, multi-part, and analytical queries; efficiently summarizing large volumes of frequently updated enterprise data; incorporating auto-ML capabilities to optimize various RAG control points automatically; and creating more robust evaluation frameworks for assessing subjective responses and conversations.

\bibliographystyle{acm}
\bibliography{sample-base}

\end{document}